\PassOptionsToPackage{hidelinks}{hyperref}

\PassOptionsToPackage{numbers,compress}{natbib}
\documentclass{article}

\usepackage[preprint]{neurips_2026}

\usepackage[utf8]{inputenc} 
\usepackage[T1]{fontenc}    
\usepackage{hyperref}       
\usepackage{url}            
\usepackage{booktabs}       
\usepackage{array}          
\usepackage{amsfonts}       
\usepackage{nicefrac}       
\usepackage[protrusion=true,expansion=false]{microtype}
\usepackage{xcolor}         

\usepackage{amsmath,amssymb}
\usepackage{enumitem}
\usepackage{graphicx}
\usepackage{float}

\makeatletter
\newcommand{\ay@sep}{}
\newcommand{\ay@paren}[1]{%
  \let\ay@sep\@empty
  \@for\ay@key:=#1\do{%
    \ay@sep
    \expandafter\citeauthor\expandafter{\ay@key},\space
    \expandafter\citeyear\expandafter{\ay@key}%
    \def\ay@sep{; }%
  }%
}
\RenewDocumentCommand{\citep}{oom}{%
  \IfNoValueTF{#2}%
    {\IfNoValueTF{#1}%
      {(\ay@paren{#3})}%
      {(\ay@paren{#3}, #1)}}%
    {(#1~\ay@paren{#3}, #2)}%
}
\makeatother

\makeatletter
\providecommand{\ct@sep}{}
\RenewDocumentCommand{\citet}{oom}{%
  \let\ct@sep\@empty
  \@for\ct@key:=#3\do{%
    \ct@sep
    \expandafter\citeauthor\expandafter{\ct@key}~[\expandafter\citenum\expandafter{\ct@key}]%
    \def\ct@sep{; }%
  }%
}
\makeatother

\title{Evaluating Type-1 Diabetes Digital Twins by Decision Quality}

\author{%
  Victor Li$^{1}$ \quad Owen Tucker$^{1}$ \quad Michael S.~Hughes$^{2}$ \\
  \bfseries Temiloluwa Prioleau$^{1}$ \quad Shengpu Tang$^{1}$ \\
  $^{1}$Department of Computer Science and $^{2}$School of Medicine, Emory University \\
  \texttt{\{victor.li, owen.tucker, michael.s.hughes,} \\
  \texttt{temiloluwa.prioleau, shengpu.tang\}@emory.edu} \\
}

\begin{document}
\maketitle
\begin{abstract}
A digital twin (DT) for type 1 diabetes (T1D) can predict how an individual's blood glucose will respond to different insulin treatments without actually trying them. For each treatment, downstream statistics of predicted glucose trajectories, such as glycemic risk, can be computed to guide clinical decisions. However, the quality of DTs is usually measured by the accuracy of their predicted glucose trajectories, which may not reflect the quality of the downstream treatment decisions they inform. In this work, we directly evaluate DTs by this decision quality on downstream treatment ranking, i.e., whether a DT predicts glucose trajectories that can be used to rank candidate treatments as the real patient would. We present an evaluation protocol over 30 virtual patients from the FDA-accepted UVA/Padova model, each simulated under 21 different basal-bolus treatments. For each patient, a DT is fitted under one treatment and then produces glucose trajectories for the remaining 20 treatments. Our results show that prediction accuracy and decision quality disagree. A baseline that models the average patient, never seeing the individual's glucose data, has the lowest prediction accuracy of all six methods (RMSE: 63.6 mg/dL) but ranks treatments with the second-highest mean Spearman (0.766), while a black-box neural network that predicts glucose nearly twice as accurately (RMSE: 32.9 mg/dL) ranks them no better than a random ordering of the 20 treatments (0.049 Spearman). Therefore, selecting a DT based on prediction accuracy alone can prefer one that recommends treatments at chance, and decision quality must be evaluated directly rather than assumed to follow from prediction accuracy.
\end{abstract}

\section{Introduction}
Type 1 diabetes (T1D) is an autoimmune condition in which the body stops producing insulin, requiring lifelong insulin therapy and frequent monitoring of blood glucose~\cite{dimeglio2018type1}. Management of glucose must therefore replicate the control loop that a healthy pancreas runs, usually through a constant basal dose covering fasting needs, meal boluses proportional to carbohydrate content, and correction boluses proportional to how far glucose sits above target. In practice, this basal-bolus treatment is specified by parameters such as the basal rate, carbohydrate ratio, and correction factor~\cite{schmidt2014bolus}. For a fixed insulin formulation and delivery method, it is these patient-specific parameters that constitute a treatment in our work. Choosing the treatment well is both difficult and extremely consequential to the patient's health. The right setting differs across patients and changes with physical activity, illness, and time of day~\cite{dimeglio2018type1,hinshaw2013diurnal,schmidt2014bolus}. Continuous glucose monitoring (CGM) has improved matters considerably, but treatment selection still relies heavily on patient-specific parameters identified through an empirical loop: make an adjustment, observe prospective glucose values, and adjust again. 

A digital twin (DT) is a computational model that acts as the virtual stand-in for the patient~\cite{katsoulakis2024digitaltwins}. For T1D, it is a personalized model that predicts how the patient's blood glucose will respond to insulin and carbohydrate intake, fit from their own glucose, insulin, and meal trace recorded under the treatment they are currently on~\cite{cappon2023replaybg,hoang2025realtime}. Rather than testing candidate treatments directly on the patient, the DT generates simulated glucose trajectories by applying these candidate treatments to itself. Each trajectory is then scored by glycemic risk, which summarizes how far and how long glucose strays from the target range~\cite{magni2007model}. Ranking the candidates by this score can give clinicians a basis for choosing among treatment options. In other words, the DT exists to answer the counterfactual question: what would happen under new treatments?

Published works evaluate DTs for T1D by prediction accuracy: how closely the predicted glucose trajectory matches the observed one, measured by root mean squared error (RMSE) or mean absolute relative difference (MARD)~\cite{cappon2023replaybg,hoang2025realtime}. However, neither measures what a DT is built to do. What matters is decision quality: whether the trajectories a DT generates rank the candidate treatments in the right order~\cite{amad2026dt2}. In fact, clinical targets are set on glycemic outcomes such as time in range~\cite{battelino2019tir}, so the question is whether an adjustment moves the patient in the right direction and how much better the best alternative would have been, rather than the pointwise accuracy of a glucose forecast. High prediction accuracy gives no guarantee that a DT's ranking of the candidate treatments matches the patient's own ranking~\cite{amad2026dt2}.

In this work, we propose an evaluation protocol that scores a DT on the decisions it supports rather than solely on the trajectories it produces, and implement it as a benchmark on three published methods~\cite{cappon2023replaybg,hoang2025realtime,hajizadeh2018plasma} and three proposed baselines. The protocol is designed for DT developers rather than for clinicians, giving evidence that a DT supports the treatment decisions a clinician would make with it and exposing DTs that predict glucose accurately while ranking treatments inaccurately. Each DT is fit to a patient's glucose trajectory under a single baseline treatment and predicts glucose under 20 counterfactual treatments. A glycemic risk kernel scores each predicted trajectory, yielding the DT's ranking over the 20 candidates. We utilize a cohort of virtual patients, so that the true ranking is obtained by running every candidate on the same patient in the simulator and applying the same kernel. Decision quality is the agreement between these rankings, measured by Spearman correlation and regret, and is reported alongside prediction accuracy for every method. Our contributions are:
\begin{itemize}[leftmargin=*,itemsep=2pt,topsep=2pt]
    \item An evaluation protocol that scores a DT for T1D by how well its ranking of candidate treatments agrees with the patient's true ranking.
    \item A benchmark implementing this protocol on 30 UVA/Padova virtual patients and 20 counterfactual treatments, with three published DT methods, each kept consistent with its original implementation, and three baselines.
    \item Evidence that the two families of metrics disagree. Methods significantly separated on prediction accuracy are not separated on decision quality, a baseline that never reads the patient's CGM has the second-highest mean Spearman, and a black-box baseline that predicts glucose nearly twice as accurately ranks treatments at chance.
\end{itemize}

\section{Related Work} \label{sec:related}
\paragraph{Digital twins for T1D.} Personalized models of a patient's glucose response are identified from their own CGM, insulin and meal record. Physiological models are fit by Bayesian MCMC in ReplayBG~\cite{cappon2023replaybg}, by neural posterior estimation~\cite{hoang2025realtime}, or by recursive filtering~\cite{hajizadeh2018plasma}, while black-box models learn the map from inputs to glucose directly from data. \citet{cappon2023blackbox} compare individualized black-box predictors with a physiological model and find the black-box approaches preferable for glucose prediction, which is why our benchmark includes two black-box baselines. These works validate a model by how closely its predictions match held-out measurements. \citet{delfavero2012glucosemetric} argue that pointwise error should weigh clinically consequential mistakes more heavily and propose a glucose-specific metric for assessing predictors, but it remains a property of a single predicted trajectory rather than of the decision a set of trajectories supports.

\paragraph{Decision-aware learning.} Decision-focused learning trains a predictor through the downstream optimization so that its errors are weighted by their effect on the decision~\cite{elmachtoub2022spo,wilder2019melding}, starting from the observation that the most accurate predictor need not yield the best decision. \citet{amad2026dt2} carry this argument to digital twins, targeting the twin at the decisions it informs. These works change how a model is trained. We hold training fixed at each method's published procedure and change only how the resulting DT is evaluated, so the protocol applies to existing T1D DT methods without retraining them.

\section{Evaluation Protocol} \label{sec:protocol}

\subsection{Setup}
Let a treatment $\pi$ be a setting of the parameters that determine a patient's insulin delivery, let $\pi_0$ be the patient's own baseline treatment, and let $\Pi$ be the set of candidate treatments, where $\pi_0\notin\Pi$. Running treatment $\pi$ on patient $p$ produces trajectory $\text{IG}^{p,\pi}$, which a glycemic risk kernel maps to a scalar reward $R(\text{IG}^{p,\pi})$. The output trajectory is defined to be noise-free interstitial glucose (IG), rather than noisy CGM, because sensor noise describes the device rather than the patient. 
Ranking $\Pi$ by $R(\text{IG}^{p,\pi})$ gives the patient's ranking over the candidate treatments. With a simulated patient cohort, we run every $\pi \in \Pi$ on the same patient under the same conditions.

\subsection{Digital Twins}
A DT method $\varphi \in \mathcal{F}$ maps a patient's noisy CGM, recorded under the baseline treatment $\pi_0$, to a DT $\tau^p \in \mathcal{T}$, so $\varphi:\mathrm{CGM}^{p,\pi_0} \longmapsto \tau^p$. A DT is itself a function, mapping the carbohydrate rate $d$ and the insulin rate $u^\pi$ delivered under some treatment $\pi$ to a predicted noise-free IG trajectory, $\tau^p:(d,\ u^\pi) \longmapsto \widehat{\mathrm{IG}}^{p,\pi}$. Using data recorded only under $\pi_0$, each DT is fit once and used to simulate glucose trajectories for every candidate treatment $\pi \in \Pi$. These two definitions are also the only assumptions required for defining a DT method. They do not need to have specific states or underlying physiological models, so mechanistic and black-box DT methods are scored identically.

\subsection{Metrics}
We evaluate a DT on the two families of metrics defined in Table~\ref{tab:metrics-table}, all computed per patient and test day on the noise-free IG. Prediction accuracy metrics are the classic RMSE and MARD scores between predicted and true trajectories, averaged over $\Pi$. Decision quality metrics score the ordering a DT induces over $\Pi$ when its predicted trajectories are passed through the glycemic risk kernel $R$. Spearman is the rank correlation between the DT's ordering and the patient's true ordering. Regret scores only the top treatment that the DT recommends as the true reward it gives up against the true best treatment, reported in both reward units and normalized across $\Pi$ so that patients with easy and hard decision problems are comparable. Which decision quality metric to prefer depends on how the DT is deployed. Regret is better when a single recommended treatment goes straight to the patient, while Spearman is better for global ranking alignment, and we choose Spearman as the primary metric.

\begin{table}[h]
\caption{Metric definitions. $T$ is the number of samples and $\widehat{\pi} = \arg\max_{\pi \in \Pi} R(\widehat{\text{IG}}^{p,\pi})$.}
\label{tab:metrics-table}
\centering
\small
\setlength{\tabcolsep}{2.5pt}
\begin{tabular}{@{}c|c@{}}
\toprule
\textbf{Prediction accuracy} & \textbf{Decision quality} \\
\cmidrule(r){1-1}\cmidrule(l){2-2}
\begin{tabular}[t]{@{}ll@{}}
RMSE & $\dfrac{1}{|\Pi|}\sum\limits_{\pi\in\Pi}\sqrt{\dfrac{1}{T}\sum\limits_{k=1}^{T}\Big(\widehat{\mathrm{IG}}^{p,\pi}_k-\mathrm{IG}^{p,\pi}_k\Big)^{2}}$ \\[1.8em]
MARD & $\dfrac{100}{|\Pi|}\sum\limits_{\pi\in\Pi}\dfrac{1}{T}\sum\limits_{k=1}^{T}\dfrac{\big|\widehat{\mathrm{IG}}^{p,\pi}_k-\mathrm{IG}^{p,\pi}_k\big|}{\mathrm{IG}^{p,\pi}_k}$ \\
\end{tabular}
&
\begin{tabular}[t]{@{}ll@{}}
Spearman & $\mathrm{corr}\big(\mathrm{rank}(R(\mathrm{IG}^{p,\pi})),\mathrm{rank}(R(\widehat{\mathrm{IG}}^{p,\pi}))\big)_{\pi\in\Pi}$ \\[0.4em]
Regret & $\max\limits_{\pi\in\Pi} R(\mathrm{IG}^{p,\pi}) - R(\mathrm{IG}^{p,\widehat{\pi}})$ \\[0.4em]
Regret Norm & $\text{Regret}\big/\big(\max\limits_{\pi}R(\mathrm{IG}^{p,\pi})-\min\limits_{\pi}R(\mathrm{IG}^{p,\pi})\big)$ \\
\end{tabular} \\
\bottomrule
\end{tabular}
\end{table}

\subsection{Statistical Analysis}
Metrics are first averaged across test days within a patient and compared to remove the varying difficulty of treatment selection across the cohort. We follow Dem\v{s}ar's protocol~\cite{demsar2006statistical} and run an omnibus Friedman test per metric with $\alpha=0.05$, ranking the methods per patient. Since the Friedman test does not identify which pair produced the rejection, we run pairwise Wilcoxon signed-rank tests with a Holm $\alpha$ correction~\cite{benavoli2016should}. We present the outcome as critical difference diagrams~\cite{demsar2006statistical, ismailfawaz2019deep} for the primary metric of each family, RMSE and Spearman. Each method is placed at its mean rank across patients, joined by a bar when the Holm-corrected pairwise test does not separate them.

\section{Benchmark Design} \label{sec:benchmark}
Our benchmark evaluates DT methods for T1D following the evaluation protocol defined above. Every design choice follows a \textit{simglucose} default~\cite{xie2018simglucose} or a published protocol unless marked as ours, and Table~\ref{tab:benchmark-design} attributes each choice to its source. Each method is kept consistent with its original implementation.

\begin{table}[t]
\caption{Benchmark summary. The last column names the source of each choice, either a \textit{simglucose} default, a published protocol, or our own design.}
\centering
\small
\begin{tabular}{@{}>{\raggedright\arraybackslash}p{2.4cm} >{\raggedright\arraybackslash}p{6.0cm} >{\raggedright\arraybackslash}p{4.6cm}@{}}
\toprule
Component & Choice & Source \\
\midrule
Simulator & \textit{simglucose} (UVA/Padova), 1 min dopri5 integration & \textit{simglucose} default~\cite{xie2018simglucose} \\
CGM & Breton--Kovatchev noise model, 5 min interval $\Rightarrow$ 288 samples per 24 h day & \textit{simglucose} GuardianRT sensor~\cite{xie2018simglucose} \\
Subjects & 30 built-in virtual patients: 10 adults, 10 adolescents, 10 children & \textit{simglucose} default~\cite{xie2018simglucose} \\
Meals & \texttt{RandomScenario}, one independent draw per day (Appendix~\ref{appendix:scenario}) & \textit{simglucose} default~\cite{xie2018simglucose} \\
Bolus formula & BBController meal bolus with a correction above 150 mg/dL (Equation~\ref{eq:bolus}) & \textit{simglucose} default~\cite{xie2018simglucose} \\
Correction bolus & computed from noise-free IG rather than CGM, so delivered insulin is deterministic given the patient, meals and treatment & ours; ReplayBG~\cite{cappon2023replaybg} instead replays a fixed insulin record \\
Candidate treatments & 20 counterfactuals: bolus or basal scaled by $-50\%$ to $+50\%$ in $10\%$ steps, each beginning at the window open & ReplayBG Scenarios 1 and 3~\cite{cappon2023replaybg}; the window start is ours \\
Risk kernel & negative total Magni risk over the scored window, glucose clipped to $[40,400]$ mg/dL (Equation~\ref{eq:magni-risk}) & \citet{magni2007model}; clip range from~\cite{fda2018den170088} \\
Identification window & 22 h at 5 min (264 samples) for MCMC, SBI and UKF; all 30 days for Linear and Neural; none for Prior & \citet{hoang2025realtime}, after the ReplayBG implementation; UKF matched to MCMC; baselines ours \\
Priors & the published ReplayBG prior for MCMC, SBI, and Prior & \citet{cappon2023replaybg} \\
Training data & 30 independent identification days under the training treatment & ours \\
Test data & 10 independent identification days $\times$ 20 treatments & ours \\
\bottomrule
\end{tabular}
\label{tab:benchmark-design}
\end{table}

\subsection{Simulator and Patient Cohort}
We use the \textit{simglucose} simulator, an open-source Python implementation of the FDA-accepted UVA/Padova T1D model~\cite{dallaman2014uvapadova, xie2018simglucose}. It has 13 states and about 30 patient-specific parameters and is treated as a black box, mirroring real clinical scenarios where the true dynamics of blood glucose are unknown. We extract noisy CGM for training and noise-free IG for ground truth, so measurement noise is present only in training, not in evaluation. Our patient cohort consists of the 30 virtual patients that ship with \textit{simglucose}: 10 adults, 10 adolescents, and 10 children~\cite{xie2018simglucose}. Each patient carries its own parameters and baseline treatment, including basal rate, carbohydrate ratio, and correction factor.

\subsection{Candidate Treatments and Risk Kernel}
Each candidate treatment is a pair of multipliers $\pi = (r_{\text{bolus}}, r_{\text{basal}})$ on the patient's baseline basal-bolus treatment, where the baseline is $\pi_0 = (1,1)$. The meal bolus follows \textit{simglucose}'s BBController formula~\cite{xie2018simglucose}, given in Equation~\ref{eq:bolus}, where $d(t)$ is the carbohydrate rate, $\Delta$ is the sample interval, CR is the patient's carbohydrate ratio, and CF is the patient's correction factor. Boluses only fire during meals, with a correction added when IG is above 150 mg/dL, and the correction term only acts upwards. $r_{\text{bolus}}$ multiplies the sum of the meal bolus and correction bolus, and $r_{\text{basal}}$ multiplies the basal rate.
\begin{equation}
    \text{bolus}(t) = r_{\text{bolus}} \cdot \Big[\underbrace{\frac{d(t)\Delta}{\text{CR}}}_{\text{carb}}+\underbrace{\mathbf{1}[d(t)>0]\cdot\mathbf{1}[\text{IG}(t)>150]\cdot \frac{\text{IG}(t)-140}{\text{CF}}}_{\text{correction}}\Big]
    \label{eq:bolus}
\end{equation}
The original BBController computes the correction term from the CGM reading, which lets measurement noise enter the delivered insulin and, from there, the trajectory and the true ranking. \citet{cappon2023replaybg} scale the originally delivered doses, so every counterfactual replays a fixed insulin record, which removes the feedback loop rather than de-noising it. We instead keep the loop closed and compute the correction from the noise-free IG, so the true ranking is a deterministic property of the patient, the meals and the treatment, regardless of the sensor seed. The set of candidate treatments also follows~\citet{cappon2023replaybg}. Either the meal bolus or the basal infusion rate is scaled by $-50\%$ to $+50\%$ in steps of $10\%$, excluding $0\%$, while the other is left unchanged. Furthermore, each candidate treatment begins at the start of the identification window rather than at the start of the simulation. Every run spends its first five hours on the patient's own baseline treatment $\pi_0$ and switches to the candidate at 5am, so every treatment enters the scored window from the same state.

We instantiate the reward $R$ as the negative total Magni risk~\cite{magni2007model} in Equation~\ref{eq:magni-risk}, where $g_t$ is the glucose at sample $t$.
\begin{equation}
  f(g) = 3.550313\left[(\ln g)^{0.835334} - 3.793357\right],~
  \mathrm{risk}(g) = 10\,f(g)^2,~
  R = -\sum_t \mathrm{risk}(g_t)
  \label{eq:magni-risk}
\end{equation}
 The kernel is near zero across the euglycemic range and grows in both directions, but asymmetrically: its constants are fixed by $\text{risk}(70) = \text{risk}(280) = 25$ and $\text{risk}(50) = \text{risk}(400)$~\cite{magni2007model}. This is what separates the reward from RMSE and MARD, which weigh a mg/dL of error the same wherever it falls. The reward of a trajectory is the negative sum of that risk over every sample in the scored window. Glucose is clipped to $[40,400]$ mg/dL before the kernel is applied, reflecting the reportable range of the Dexcom G6 CGM~\cite{fda2018den170088}.

\subsection{Scenario and Data}
Meals are drawn from \textit{simglucose}'s default RandomScenario: three main meals at probability $0.95$ and three snacks at probability $0.30$, yielding 1 to 6 eating occasions and $196 \pm 30$~g of carbohydrate intake per day. Each identification day covers 24 hours starting from 5am and CGM is sampled every 5 minutes. Appendix~\ref{appendix:scenario} provides a listing of all parameters used. Each scenario is drawn independently and cached, so all 21 treatments and benchmarked methods see the same meals on the same day. Differences between treatments are therefore never confounded by noise in what the patient ate.

Training data consists of 30 independent identification days under the training treatment. We cap it at 30 days, about one month of CGM, as a realistic amount of recent data for a single patient and more than twice the 14 days the international consensus recommends for assessing glycemic control~\cite{battelino2019tir}. We first create 29 hour trajectories, each with an independent scenario draw. The first 5 hours of each trajectory are dropped and never scored or handed to a method: \textit{simglucose} truncates breakfast times to 5:00am--9:00am, so 5am is the earliest instant at which any meal can be drawn. No method is required to use all 30 days, and each uses the amount its original implementation prescribes. Evaluation data consists of 10 additional independent identification days, each starting from the baseline treatment $\pi_0$ until 5am and then switching to the candidate treatment afterwards.

\subsection{Methods} \label{sec:methods}
We evaluate six DT methods: three published DT methods for T1D, all run here on the same simplified, nine-state, eight-parameter physiological ordinary differential equation (ODE) model of \citet{cappon2023replaybg}, and three baselines we propose. Each method uses the amount of training data its original implementation prescribes, so the data requirement is part of the method being evaluated. MCMC and SBI identify from a single 22 hour window (264 CGM samples), the setting \citet{hoang2025realtime} adopt from the ReplayBG implementation and the exact input of their network, and UKF is matched to the same window. Extending them to all 30 days is computationally expensive as every MCMC likelihood evaluation simulates the ODE over the whole window. The single-day requirement is therefore a property, and a limitation, of these methods rather than of the benchmark. The Linear and Neural baselines have no published setting, so we give them all 30 training days (8640 samples). The Prior baseline uses no training data.

\begin{itemize}[leftmargin=*,itemsep=3pt,topsep=2pt]
    \item \textbf{Markov chain Monte Carlo (MCMC)}~\cite{cappon2023replaybg}. The ReplayBG method: Bayesian identification of a nine-state, eight-parameter physiological ODE by adaptive single-component Metropolis--Hastings, followed by a multivariate $t$-copula over the marginals. It is identified from a single 22 hour window and replays candidate treatments from an assumed steady state.
    \item \textbf{Simulation-based inference (SBI)}~\cite{hoang2025realtime}. A masked autoregressive flow trained by neural posterior estimation to map a CGM trace directly to a posterior over the same ODE model, extended to infer the nine initial conditions jointly. The flow conditions on CGM alone, so the insulin and meal record enters only through its training set, which the original paper generates under one fixed profile. Each patient's record differs, so we train a fresh network per patient: 5,000 parameter-trace pairs are simulated from the published prior under that patient's own 22 hour record, the flow is fit to them once, and it is then conditioned on the observed CGM to draw 1,000 posterior samples. Nothing is shared across patients, so the benchmark evaluates the estimator rather than the cross-patient amortization that is the original paper's headline; its own future work targets the same limitation by training on varied meal and insulin profiles.
    \item \textbf{Unscented Kalman filter (UKF)}~\cite{hajizadeh2018plasma}. In T1D, the UKF has been used to estimate plasma insulin concentration under the Hovorka model, as a real-time estimator for artificial pancreas systems. We apply the same joint state-parameter filtering idea to the simplified ODE and estimate glucose rather than plasma insulin. The filter runs over the recorded inputs of the same 22 hour window, and candidate treatments are then replayed through the same path MCMC uses.
    \item \textbf{Linear response model (Linear)}. A ridge regression of CGM on lagged insulin and carbohydrate rates, with four hours of insulin memory and three hours of carbohydrate memory. It has no physiological model, no states, and no parameter that means anything. It regresses on the input record alone, with no autoregressive CGM terms, because a DT must generate a full day under a treatment whose CGM is never observed.
    \item \textbf{Neural response model (Neural)}. A multilayer perceptron with two hidden layers of 64 units over the same lagged features, training days and covariate as Linear, so the pair differs only in model complexity and Linear's behavior cannot be attributed to linearity. Both are black-box models, the family that outperformed an individualized physiological model at 30 to 60 minute glucose forecasting~\cite{cappon2023blackbox}.
    \item \textbf{Average patient (Prior)}. The MCMC pipeline with the identification step removed: candidate treatments are replayed over an ensemble drawn from the published ReplayBG prior, so its identification window is empty and the DT reads no patient CGM at all. It shares the ODE, steady-state initial condition, batched replay and ensemble median with MCMC, so it ablates exactly one input. It still uses the patient's basal insulin and body weight, without which the model cannot be instantiated, so it isolates CGM-based identification rather than all patient information.
\end{itemize}

\section{Results} \label{sec:results}
\begin{table}
\caption{Results over the full cohort, ordered by RMSE. Parenthesized numbers are ranks on each family's primary metric, and $\Delta$ rank is the move from the accuracy rank to the decision rank, positive meaning better on decisions than on accuracy.}
\label{tab:results}
\centering
\small
\setlength{\tabcolsep}{4pt}
\begin{tabular}{@{}lrr|rrrr@{}}
\toprule
& \multicolumn{2}{c|}{\textbf{Prediction Accuracy}} & \multicolumn{4}{c}{\textbf{Decision Quality}} \\
\cmidrule(r){2-3}\cmidrule(l){4-7}
Method & RMSE (mg/dL)& MARD (\%) & Spearman & Regret & Regret Norm & $\Delta$ rank \\
\midrule
MCMC & $23.0 \pm 9.1$~(1) & $21.0 \pm 16.2$ & $0.861 \pm 0.111$~(1) & $370 \pm 470$ & $0.083 \pm 0.113$ & $0$ \\
SBI & $29.2 \pm 11.3$~(2) & $32.4 \pm 31.5$ & $0.728 \pm 0.368$~(3) & $584 \pm 858$ & $0.137 \pm 0.208$ & $-1$ \\
Neural & $32.9 \pm 9.0$~(3) & $31.5 \pm 17.6$ & $0.049 \pm 0.466$~(5) & $3222 \pm 2888$ & $0.541 \pm 0.327$ & $\mathbf{-2}$ \\
UKF & $35.6 \pm 20.6$~(4) & $42.5 \pm 55.3$ & $0.492 \pm 0.614$~(4) & $1297 \pm 1979$ & $0.215 \pm 0.278$ & $0$ \\
Linear & $36.6 \pm 8.7$~(5) & $36.8 \pm 23.5$ & $-0.207 \pm 0.524$~(6) & $4068 \pm 3801$ & $0.654 \pm 0.342$ & $-1$ \\
Prior & $63.6 \pm 31.4$~(6) & $52.7 \pm 38.2$ & $0.766 \pm 0.156$~(2) & $876 \pm 1324$ & $0.174 \pm 0.151$ & $\mathbf{+4}$ \\
\bottomrule
\end{tabular}
\end{table}

We run the evaluation protocol on all six methods over one shared set of cached trajectories. Table~\ref{tab:results} reports every metric over the full cohort as the mean $\pm$ standard deviation across the 30 virtual patients. The omnibus Friedman test rejects on all five metrics (Table~\ref{tab:omnibus}), and the Holm-corrected pairwise Wilcoxon tests that follow it are given in full in Appendix~\ref{appendix:statistical-tests}. Figure~\ref{fig:cd-diagrams} plots the critical difference diagrams for the primary metric of each family, RMSE and Spearman. In the pairwise tests, four pairs of methods show the three ways in which the two families can disagree: MCMC versus SBI and SBI versus Prior are separated on RMSE but not on Spearman, UKF versus Neural is separated on Spearman but not on RMSE, and Neural versus Prior is separated on both, in opposite directions.

\begin{table}[t]
\caption{Omnibus Friedman test. Mean within-patient rank ($n=30$ patients, $k=6$ methods, rank 1 = best) and the omnibus Friedman $p$ per metric.}
\label{tab:omnibus}
\centering
\small
\begin{tabular}{lrrrrrrr}
\toprule
Metric & MCMC & SBI & UKF & Linear & Neural & Prior & Friedman $p$ \\
\midrule
RMSE & 1.47 & 2.67 & 3.37 & 4.60 & 3.37 & 5.53 & $<10^{-4}$ \\
MARD & 1.57 & 2.83 & 3.27 & 4.63 & 3.33 & 5.37 & $<10^{-4}$ \\
Spearman & 1.87 & 2.57 & 3.37 & 5.27 & 4.83 & 3.10 & $<10^{-4}$ \\
Regret & 2.10 & 2.77 & 3.03 & 4.88 & 4.88 & 3.33 & $<10^{-4}$ \\
Regret Norm & 2.07 & 2.77 & 3.00 & 4.92 & 4.92 & 3.33 & $<10^{-4}$ \\
\bottomrule
\end{tabular}
\end{table}

\begin{figure}
\centering
\includegraphics[width=\linewidth]{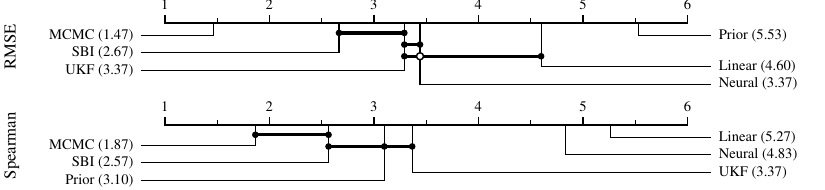}
\caption{Critical difference diagrams for RMSE and Spearman. Each method sits at its mean rank over the 30 patients (1 is best). A bar joins methods that the Holm-corrected pairwise Wilcoxon tests of Table~\ref{tab:posthoc} cannot separate at $\alpha=0.05$. A filled circle marks a member of the group, and an open circle marks a method the bar passes over but does not include.}
\label{fig:cd-diagrams}
\end{figure}

The two families of metrics do not order the methods the same way. Ordering Table~\ref{tab:results} by RMSE leaves the Spearman order at 1, 3, 5, 4, 6, 2. Prior sits at the bottom of the prediction accuracy panel with 63.6 mg/dL RMSE, while also near the top of the decision quality panel with 0.766 Spearman, which we cannot distinguish from SBI's 0.728 ($p = 0.3931$). Since Prior carries the population's average physiology, this suggests that rankings over these 20 treatments are largely set by the direction of the average dose response and that personalization refines the ordering rather than defining it. Without reading patient CGM at all, a prior-based DT can therefore be the worst trajectory predictor by construction and still be among the best at the decision task DTs are built for. Furthermore, MCMC and SBI are statistically different on the prediction accuracy metric, RMSE difference of 6.25 mg/dL ($p = 0.0002$), while we cannot distinguish them on the decision quality metric, 0.133 Spearman difference ($p = 0.1706$). A user choosing a DT method by prediction accuracy might confidently take MCMC over SBI, even though we cannot distinguish the two in decision quality. Likewise, the Neural and Prior baselines invert between the two families: Neural is 30.65 mg/dL better on RMSE and 0.717 worse on Spearman ($p < 10^{-4}$ for both), so choosing between them on prediction accuracy alone leads to a DT that is essentially random in its downstream recommendations.

Interestingly, the two black-box baselines do not behave as expected. Neural saw a significant improvement over Linear on RMSE ($p <10^{-4}$), but Spearman moved from inverting the ranking to carrying no information about it at all. At 0.049 Spearman, the Neural baseline shows that a DT whose RMSE we cannot distinguish from that of the published method UKF ($p = 0.9515$) can carry no decision quality at all. We also cannot distinguish UKF from Linear on RMSE ($p = 0.5602$), so prediction accuracy separates it from neither black-box baseline while decision quality separates it from both ($p < 10^{-4}$ against Linear, $p = 0.0061$ against Neural).

\paragraph{Sensitivity.} To check that these conclusions do not hinge on individual patients, we repeated the entire analysis 30 times, each time leaving out one patient and recomputing every test on the remaining 29. Every comparison stated in this section holds in all 30 repetitions. We also tested leaving out a whole age group of 10 patients, under which the MCMC versus SBI, SBI versus Prior, and Neural versus Prior comparisons still hold, while some finer orderings among the other methods change.

\section{Discussion} \label{sec:discussion}
The decision quality metrics make room for this dissociation by construction. Regret evaluates $R$ on true trajectories in both of its terms, so a DT's prediction enters only through which treatment $\widehat{\pi}$ it selects. A DT whose predicted glucose is uniformly biased is therefore not penalized for the bias unless the bias changes its choice, while RMSE and MARD charge for it at every sample. Spearman extends the same property to the whole ranking: any prediction error that preserves the order of the predicted rewards costs nothing. Prior is the extreme case. It is the least accurate predictor by a wide margin, yet its ranking agrees with the patient's at 0.766 Spearman, so most of its error leaves the order of the predicted rewards intact.

The intended benefit of our work is to lower the chance of deploying a DT that predicts glucose accurately and still recommends the wrong insulin dose. The corresponding risk is that a strong score on this benchmark is mistaken for clinical validation, which it is not. All code is available at \url{https://github.com/Veeeeeee7/T1D-Digital-Twin-Benchmark}.

\subsection{Limitations} \label{sec:limitations}
Glucose is clipped to $[40,400]$ mg/dL so that regret is not dominated by a single deep excursion, which means 39 and 15 mg/dL score identically. Of the 6000 ground-truth runs, 205 ($3.4\%$), concentrated in 8 patients and almost entirely on the increased-insulin treatments, left the simulator's safe range and were terminated early and padded at the terminal value. The ReplayBG prior comes from earlier studies and ReplayBG was assessed on virtual adults~\cite{cappon2023replaybg}, so the 20 adolescent and child subjects may carry extra prior mismatch. The candidate set moves a single dose axis at a time and contains only insulin adjustments, whereas clinical adjustments may move basal and prandial doses jointly and sometimes in opposite directions. Finally, every result is for one candidate set, one risk kernel and a 24 hour horizon, and we do not know how much of the dissociation survives a different objective such as time in range or a longer horizon.

\section{Conclusion} \label{sec:conclusion}
We proposed evaluating T1D digital twins by the treatment decisions they support. On 30 virtual patients, prediction accuracy and decision quality disagree on which of the six methods are better, and selecting a DT by prediction accuracy alone can prefer one that recommends treatments at chance, as the Neural and Prior baselines show. Decision quality must therefore be reported alongside RMSE and MARD rather than assumed to follow from them.

\bibliographystyle{plainnat}
\bibliography{ref}

\newpage
\appendix
\section{Benchmark Details} \label{appendix:benchmark-details}

\subsection{Scenario} \label{appendix:scenario}
Table~\ref{tab:scenario-table} summarizes the RandomScenario implemented in \textit{simglucose}~\cite{xie2018simglucose}.

\begin{table}[h]
\caption{Scenario Table. $p$ is the probability of a given meal or snack occurring.}
\centering
\begin{tabular}{lp{8.4cm}}
\toprule
Horizon & 24 h starting at 5am after a steady-state start at midnight \\
CGM Interval & 5 min $\Rightarrow$ 288 samples per test day \\
\midrule
\multicolumn{2}{l}{Simglucose Default RandomScenario} \\
Main Meals & breakfast / lunch / dinner at $p=0.95$, mealtimes at $\mathcal{N}$(07:00, 12:00, 18:00; $\sigma=60$ min) truncated to $[05,09]$, $[10,14]$, $[16,20]$, amounts of $\mathcal{N}$(45g, 70g, 80g; $\sigma=10$g) \\
Snacks & three at $p=0.30$, mealtimes at $\mathcal{N}$(09:30, 15:00, 21:30; $\sigma=30$ min), amounts of $\mathcal{N}$(10g; $\sigma=5$g) \\
Measured & 1--6 eating occasions per day, $196 \pm 30$ g/day \\
\bottomrule
\end{tabular}
\label{tab:scenario-table}
\end{table}

\subsection{Existing assets} \label{appendix:assets}
The simulator, the 30 virtual subjects, the Breton--Kovatchev CGM noise model and \texttt{RandomScenario} come from \textit{simglucose} v0.2.11 (MIT license)~\cite{xie2018simglucose}, an independent open-source Python implementation of the 2008 UVA/Padova simulator~\cite{kovatchev2009insilico} distributed for research use. Neural posterior estimation uses the \texttt{sbi} toolkit v0.26.1 (Apache-2.0)~\cite{tejerocantero2020sbi} on PyTorch v2.12 (BSD-3-Clause), with NumPy, SciPy and pandas (BSD-3-Clause), Gymnasium (MIT) and Matplotlib (PSF-based Matplotlib license). The ReplayBG model, its published prior and its MCMC procedure~\cite{cappon2023replaybg}, the NPE setup of~\citet{hoang2025realtime}, and the unscented Kalman filter of~\citet{hajizadeh2018plasma} are re-implemented from the published descriptions and those papers are credited as specifications.

\section{Compute Resources} \label{appendix:compute}
Every experiment runs on CPU only. All runs were submitted to an internal SLURM cluster of 64-core, 512 GB nodes, each job requesting 30 cores and 128 GB with one worker per patient, so the cohort is fit and scored concurrently. Table~\ref{tab:compute} gives the measured cost of fitting one patient and scoring all 20 candidate treatments on a single core, as the median over the 30 patients.

\begin{table}[H]
\caption{Per-patient fit-and-score cost on one core, median over the 30 subjects.}
\label{tab:compute}
\centering
\small
\begin{tabular}{lrrrrrr}
\toprule
& MCMC & SBI & UKF & Prior & Neural & Linear \\
\midrule
Cost per patient & 8 h 18 m & 245 s & 181 s & 175 s & 2.4 s & 0.1 s \\
\bottomrule
\end{tabular}
\end{table}

The five fast methods ran as a single job in 861 s of wall clock, 5.5 core-hours in total. MCMC ran separately in 8 h 30 m of wall clock, 249 core-hours. Generating the shared trajectory cache, 30 patients $\times$ 30 training days plus 10 test days $\times$ 20 treatments, costs a further 1182 s of wall clock, 9.9 core-hours, and is done once and reused by every method.

\section{Statistical Tests} \label{appendix:statistical-tests}
The omnibus Friedman tests are in Table~\ref{tab:omnibus}, and every one rejects. Table~\ref{tab:posthoc} shows the pairwise Wilcoxon signed-rank tests with Holm correction for each pair of methods on RMSE, MARD, Spearman and Regret Norm. Both tables are computed on the within-patient means, $n = 30$ patients, following the protocol of Section~\ref{sec:protocol}.

\begin{table}
\caption{Pairwise Wilcoxon signed-rank tests with Holm correction within each metric. $\Delta$ is the mean within-patient difference.}
\label{tab:posthoc}
\centering
\footnotesize
\begin{tabular}{llrrrc}
\toprule
Metric & A vs B & $\Delta$ (A$-$B) & raw $p$ & Holm $p$ & significant at $0.05$ \\
\midrule
RMSE & MCMC vs SBI & $-6.25$ & $<10^{-4}$ & $0.0002$ & yes \\
 & MCMC vs UKF & $-12.61$ & $<10^{-4}$ & $<10^{-4}$ & yes \\
 & MCMC vs Linear & $-13.58$ & $<10^{-4}$ & $<10^{-4}$ & yes \\
 & MCMC vs Neural & $-9.97$ & $<10^{-4}$ & $<10^{-4}$ & yes \\
 & MCMC vs Prior & $-40.62$ & $<10^{-4}$ & $<10^{-4}$ & yes \\
 & SBI vs UKF & $-6.36$ & $0.0262$ & $0.0787$ & no \\
 & SBI vs Linear & $-7.33$ & $<10^{-4}$ & $<10^{-4}$ & yes \\
 & SBI vs Neural & $-3.72$ & $0.0022$ & $0.0088$ & yes \\
 & SBI vs Prior & $-34.37$ & $<10^{-4}$ & $<10^{-4}$ & yes \\
 & UKF vs Linear & $-0.97$ & $0.2801$ & $0.5602$ & no \\
 & UKF vs Neural & $+2.65$ & $0.9515$ & $0.9515$ & no \\
 & UKF vs Prior & $-28.00$ & $<10^{-4}$ & $<10^{-4}$ & yes \\
 & Linear vs Neural & $+3.62$ & $<10^{-4}$ & $<10^{-4}$ & yes \\
 & Linear vs Prior & $-27.03$ & $<10^{-4}$ & $<10^{-4}$ & yes \\
 & Neural vs Prior & $-30.65$ & $<10^{-4}$ & $<10^{-4}$ & yes \\
\midrule
MARD & MCMC vs SBI & $-11.40$ & $<10^{-4}$ & $<10^{-4}$ & yes \\
 & MCMC vs UKF & $-21.51$ & $<10^{-4}$ & $<10^{-4}$ & yes \\
 & MCMC vs Linear & $-15.77$ & $<10^{-4}$ & $<10^{-4}$ & yes \\
 & MCMC vs Neural & $-10.52$ & $<10^{-4}$ & $<10^{-4}$ & yes \\
 & MCMC vs Prior & $-31.72$ & $<10^{-4}$ & $<10^{-4}$ & yes \\
 & SBI vs UKF & $-10.11$ & $0.0732$ & $0.2930$ & no \\
 & SBI vs Linear & $-4.36$ & $0.0006$ & $0.0037$ & yes \\
 & SBI vs Neural & $+0.88$ & $0.1519$ & $0.4208$ & no \\
 & SBI vs Prior & $-20.31$ & $<10^{-4}$ & $<10^{-4}$ & yes \\
 & UKF vs Linear & $+5.74$ & $0.1403$ & $0.4208$ & no \\
 & UKF vs Neural & $+10.99$ & $0.8553$ & $0.8553$ & no \\
 & UKF vs Prior & $-10.21$ & $0.0137$ & $0.0683$ & no \\
 & Linear vs Neural & $+5.25$ & $<10^{-4}$ & $<10^{-4}$ & yes \\
 & Linear vs Prior & $-15.95$ & $<10^{-4}$ & $0.0004$ & yes \\
 & Neural vs Prior & $-21.20$ & $<10^{-4}$ & $<10^{-4}$ & yes \\
\midrule
Spearman & MCMC vs SBI & $+0.133$ & $0.0427$ & $0.1706$ & no \\
 & MCMC vs UKF & $+0.369$ & $0.0030$ & $0.0149$ & yes \\
 & MCMC vs Linear & $+1.069$ & $<10^{-4}$ & $<10^{-4}$ & yes \\
 & MCMC vs Neural & $+0.812$ & $<10^{-4}$ & $<10^{-4}$ & yes \\
 & MCMC vs Prior & $+0.095$ & $0.0001$ & $0.0010$ & yes \\
 & SBI vs UKF & $+0.236$ & $0.0732$ & $0.2197$ & no \\
 & SBI vs Linear & $+0.936$ & $<10^{-4}$ & $<10^{-4}$ & yes \\
 & SBI vs Neural & $+0.679$ & $<10^{-4}$ & $<10^{-4}$ & yes \\
 & SBI vs Prior & $-0.038$ & $0.3931$ & $0.3931$ & no \\
 & UKF vs Linear & $+0.700$ & $<10^{-4}$ & $<10^{-4}$ & yes \\
 & UKF vs Neural & $+0.443$ & $0.0009$ & $0.0061$ & yes \\
 & UKF vs Prior & $-0.274$ & $0.1909$ & $0.3819$ & no \\
 & Linear vs Neural & $-0.257$ & $0.0011$ & $0.0068$ & yes \\
 & Linear vs Prior & $-0.974$ & $<10^{-4}$ & $<10^{-4}$ & yes \\
 & Neural vs Prior & $-0.717$ & $<10^{-4}$ & $<10^{-4}$ & yes \\
\midrule
Regret Norm & MCMC vs SBI & $-0.054$ & $0.0835$ & $0.4176$ & no \\
 & MCMC vs UKF & $-0.132$ & $0.0389$ & $0.2335$ & no \\
 & MCMC vs Linear & $-0.571$ & $<10^{-4}$ & $<10^{-4}$ & yes \\
 & MCMC vs Neural & $-0.458$ & $<10^{-4}$ & $<10^{-4}$ & yes \\
 & MCMC vs Prior & $-0.091$ & $0.0002$ & $0.0014$ & yes \\
 & SBI vs UKF & $-0.078$ & $0.2943$ & $0.5886$ & no \\
 & SBI vs Linear & $-0.517$ & $<10^{-4}$ & $0.0004$ & yes \\
 & SBI vs Neural & $-0.404$ & $<10^{-4}$ & $<10^{-4}$ & yes \\
 & SBI vs Prior & $-0.037$ & $0.0964$ & $0.4176$ & no \\
 & UKF vs Linear & $-0.439$ & $<10^{-4}$ & $0.0002$ & yes \\
 & UKF vs Neural & $-0.326$ & $0.0003$ & $0.0018$ & yes \\
 & UKF vs Prior & $+0.041$ & $0.7375$ & $0.7375$ & no \\
 & Linear vs Neural & $+0.113$ & $0.0974$ & $0.4176$ & no \\
 & Linear vs Prior & $+0.480$ & $<10^{-4}$ & $0.0002$ & yes \\
 & Neural vs Prior & $+0.367$ & $<10^{-4}$ & $<10^{-4}$ & yes \\
\bottomrule
\end{tabular}
\end{table}

\end{document}